\documentclass{article}

\usepackage[nonatbib, final]{neurips}
\usepackage[numbers]{natbib}

\makeatletter
\renewcommand{\@noticestring}{
  \centering
}
\makeatother

\usepackage[export]{adjustbox}
\usepackage[ruled]{algorithm2e}
\usepackage[inline, shortlabels]{enumitem}
\usepackage[T1]{fontenc}
\usepackage{hyperref}
\usepackage{microtype}
\usepackage{pifont}
\usepackage{xcolor}
\usepackage{xurl}
\usepackage{float}
\usepackage{graphicx}
\usepackage{booktabs}
\usepackage{tabularray}
\usepackage{makecell}
\usepackage{array}
\usepackage{rotating}
\usepackage{multicol}
\usepackage{multirow}
\usepackage{listings}
\usepackage{amsmath, amsfonts}
\usepackage{nicefrac}
\usepackage{subcaption}

\UseTblrLibrary{booktabs}

\makeatletter
\newcommand{\ssymbol}[1]{\@fnsymbol{#1}}
\newcommand{\romanNumeral}[1]{\expandafter\@slowromancap\romannumeral #1@}
\makeatother

\newif \ifhq
\hqtrue

\newcommand{\nameofmethod}{HunyuanImage 3.0}
\newcolumntype{C}{>{\centering\arraybackslash}p{1.1cm}}
\newcommand{\figref}[1]{Figure~\ref{#1}}
\newcommand{\tabref}[1]{Table~\ref{#1}}

\newcommand{\secref}[1]{Section~\ref{#1}}

\title{\nameofmethod{} Technical Report}

\author{Tencent Hunyuan Foundation Model Team}

\begin{document}

\maketitle

\begin{abstract}
We present \nameofmethod{}, a native multimodal model that unifies multimodal understanding and generation within an autoregressive framework, with its image generation module publicly available.
The achievement of \nameofmethod{} relies on several key components, including meticulous data curation, advanced architecture design, a native Chain-of-Thoughts schema, progressive model pre-training, aggressive model post-training, and an efficient infrastructure that enables large-scale training and inference.
With these advancements, we successfully trained a Mixture-of-Experts (MoE) model comprising over 80 billion parameters in total, with 13 billion parameters activated per token during inference, making it the largest and most powerful open-source image generative model to date.
We conducted extensive experiments and the results of automatic and human evaluation of text-image alignment and visual quality demonstrate that \nameofmethod{} rivals previous state-of-the-art models.
By releasing the code and weights of \nameofmethod{}, we aim to enable the community to explore new ideas with a state-of-the-art foundation model, fostering a dynamic and vibrant multimodal ecosystem.
All open source assets are publicly available at \href{https://github.com/Tencent-Hunyuan/HunyuanImage-3.0}{here}.
\end{abstract}

\begin{figure}[pt]
    \vspace*{-3cm}
    \hspace*{-3.2cm}
    \includegraphics[width=1.45\textwidth]{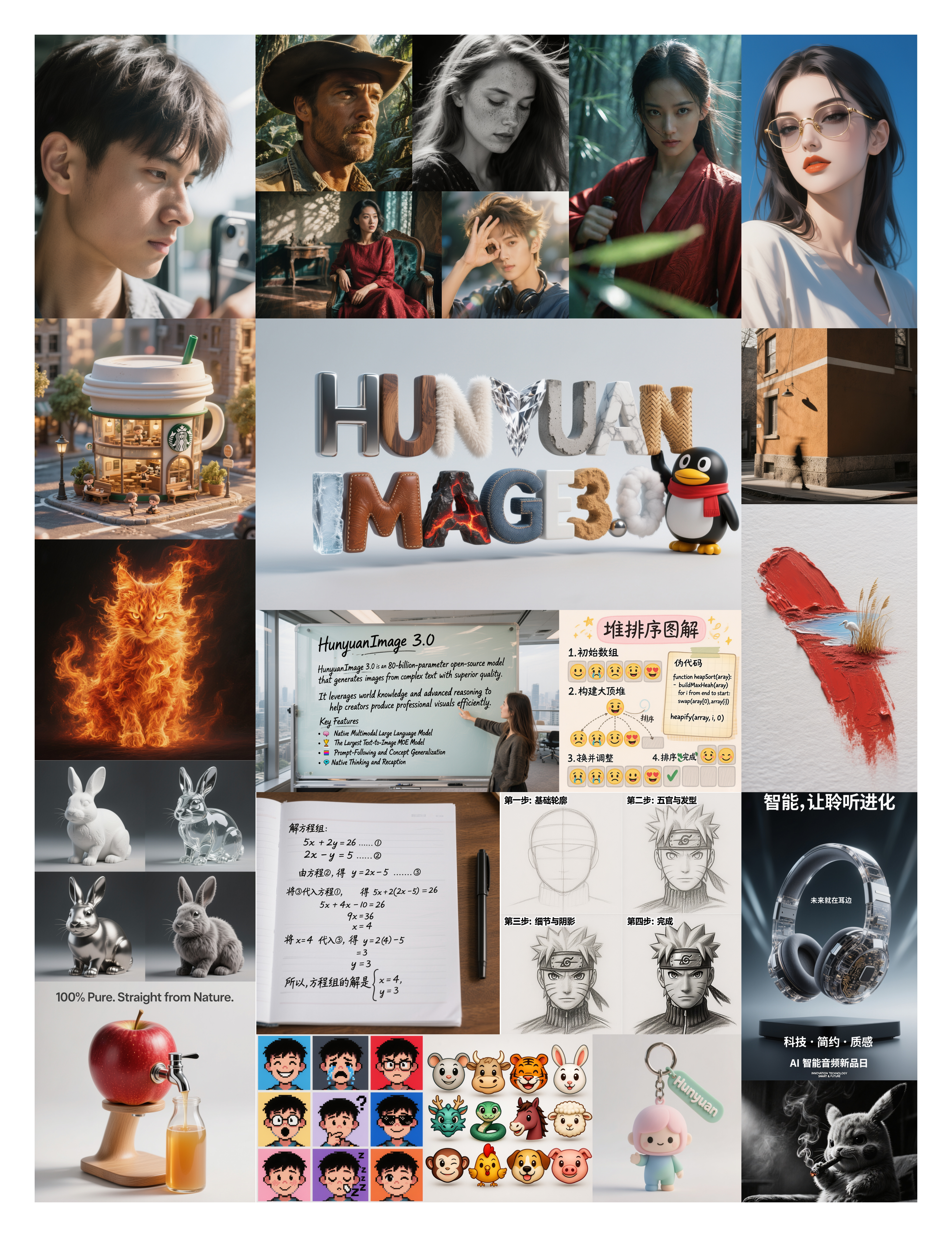}
    \vspace{-1.2cm}
    \caption{Multi-ratio text-to-image samples from \nameofmethod{}, demonstrating its powerful prompt-following, reasoning, concept generalization and text-rendering capabilities.}
    \label{fig:banner}
\end{figure}

\section{Introduction}
In recent years, image generation models have achieved remarkable progress, enabling the generation of realistic and diverse images from natural language descriptions and reference images.
Advances in deep learning architectures, particularly diffusion models~\cite{ho2020denoising, song2020denoising, dhariwal2021diffusion, song2020score, rombach2022high, karras2022elucidating, lipman2022flow, liu2022flow, zhang2022unsupervised, zhang2023shiftddpms} and transformer-based frameworks~\cite{vaswani2017attention, peebles2023scalable}, have significantly improved image fidelity and semantic alignment with input texts.
Leading models such as Seedream 4.0~\cite{Seedream-4.0}, Nano Banana~\cite{Nano-Banana}, GPT-Image~\cite{GPT-Image}, Qwen-Image~\cite{qwen_image} and HunyuanImage 2.1~\cite{HunyuanImage-2.1} have demonstrated the capability to synthesize complex scenes and artistic styles or edit image accurately, attracting widespread attention in both research and industry.
However, these state-of-the-art systems are predominantly closed-source, limiting transparency and reproducibility for the wider research community.

%

For this reason, we present \nameofmethod{}, an open-source model that achieves image generation performance comparable to, or surpassing, that of leading closed-source models.
\nameofmethod{} originates from our internally developed native multimodal model, currently with fine-tuning and post-training focused solely on image generation.
We employ Hunyuan-A13B~\cite{HunyuanA13B}, a pre-trained Mixture-of-Experts (MoE) large language model (LLM) with over 80 billion total parameters, of which 13 billion are activated per token during inference, as our base model.
The choice reconciles the need for both high model capacity and computational efficiency.
To extend the LLM to handle visual inputs for image understanding and generation, we augment it with a pre-trained vision encoder and a VAE, each equipped with a projection layer that transforms the extracted image features into a joint embedding space compatible with the LLM’s word embeddings~\cite{alayrac2022flamingo, liu2023visual, liu2024improved, zhou2024transfusion, ma2025janusflow}.
For image understanding, the LLM conditions its autoregressive next-token prediction on joint image features extracted from the vision encoder and the VAE to generate appropriate responses.
For image generation, diffusion-based image modeling on VAE image features is incorporated into the LLM in the same manner as in Transfusion~\cite{zhou2024transfusion} and JanusFlow~\cite{ma2025janusflow}.
Furthermore, the LLM-based framework enables us to incorporate Chain-of-Thought training and inference, thereby improving the performance of both image understanding and generation tasks.
After fine-tuning and post-training the pre-trained model solely on image generation tasks, we establish the image generation module of \nameofmethod{}, which currently stands as the largest and most powerful open-source image generation model.
We conduct extensive experiments on both automatic and human evaluation, the results on text-image alignment and visual quality demonstrate that \nameofmethod{} rivals previous state-of-the-art models, including Seedream 4.0~\cite{Seedream-4.0}, Nano Banana~\cite{Nano-Banana}, GPT-Image~\cite{GPT-Image} and HunyuanImage 2.1~\cite{HunyuanImage-2.1}.
By releasing the code and weights of \nameofmethod{}, we aim to enable the community to explore new ideas with a state-of-the-art foundation model, fostering a dynamic and vibrant image generation ecosystem.

This report is structured as follows.
In \textbf{\secref{sec:data}}, we introduce our data preparation techniques, including filtering and captioning models.
\textbf{\secref{sec:model}} presents detailed information about the architecture and algorithms of \nameofmethod{}.
In \textbf{\secref{sec:training}}, we discuss our training strategies and algorithms.
In \textbf{\secref{sec:performance}}, we evaluate the performance of \nameofmethod{} and compare it with state-of-the-art text-to-image generation models.

\section{Data Preparation}\label{sec:data}

\subsection{Data Filtering}
To construct a diverse, high-quality image dataset, we implemented a comprehensive three-stage filtering process on an initial pool of over 10 billion raw images. This rigorous process, which ultimately retained less than 45\% of the initial data, was designed to prioritize both semantic diversity and image quality, critical requirements for training robust generative models.

In the first stage, we addressed technical flaws by removing images with low resolution (less than 512 pixels), broken files, over-/under-exposure, and over-saturation. We also deduplicated the images according to their MD5 values. 

The second stage served as our primary data curation process, employing two types of operators: objective filters and subject-scoring operators. The objective filters were learning-based detectors for watermarks, logos, extensive text (through the hy-OCR model\footnote{to be released}), collages, prominent borders and AI-generated content (AIGC). To maintain accuracy when dealing with massive data, these detectors were trained using balanced training datasets created through stratified sampling. The proliferation of AIGC images poses a significant challenge by distorting natural data distributions and impairing model convergence. Our mitigation strategy combined an automated AIGC detection model~\cite{AIGC_1_luo2024lare,AIGC_2_zhou2025aigi,AIGC_3_chen2025dual} with the removal of all images from data sources found to have a high proportion of AI-generated content.

Our subject-scoring operators included models for image clarity and aesthetics. The clarity model provided an overall score based on an image's sharpness, noise level, and dynamic range. To ensure consistent and interpretable aesthetic scores, our artists systematically designed a criterion that considered three fundamental elements: color, light \& shadow, and composition. Based on this criterion, we build our own aesthetics model. We applied a unified threshold value to filter out unqualified images across all types, while using different threshold values for specific genres to select data for later training stages.

In the final stage, we further deduplicated data based on embedding cluster results, which removed approximately 0.5\% of the data, making our datasets more compact. To enhance semantic breadth, we strategically supplemented the filtered set with specialized datasets, including knowledge-augmented , text-related, stylized, and graphic design collections. This meticulous pipeline resulted in clean, high-quality, and diverse datasets formed with nearly 5 billion images for training advanced generative models.
Beyond the single-image corpus, we constructed a specialized dataset of over 100 million image pairs and multi-image samples specifically designed for learning interleave relationships. The image-pair subset was derived through two primary approaches: image clustering and video segment mining.

For the image clustering approach, following the analysis of over two billion image clusters, we selectively extracted pairs from the representative clusters that exhibited potential similarity. These pairs were then passed through an image-relation discrimination operator to retain only those with a significant editing relationship. To optimize the model's learning efficacy, an image complexity model~\cite{feng2023ic9600} was applied to filter out images whose constituent elements were deemed overly complex.

The video data mining pipeline began with shot boundary detection to isolate video segments corresponding to unified scenes. Camera motion classification operator was subsequently employed to exclude clips exhibiting excessive camera transformation. We further refined the selection by integrating results from object detection and semantic segmentation to isolate keyframes demonstrating canonical transformation relationships. Finally, to mitigate the detrimental effects of motion blur on model training, the selected frames underwent an additional round of filtering based on a motion blur detection operator. The resulting multi-image data is composed of interleaved data (image-text sequences) sourced from the internet and extracted frames from videos.
\subsection{Image Captioning}
\begin{figure}[h]
    \centering
    \includegraphics[width=\linewidth]{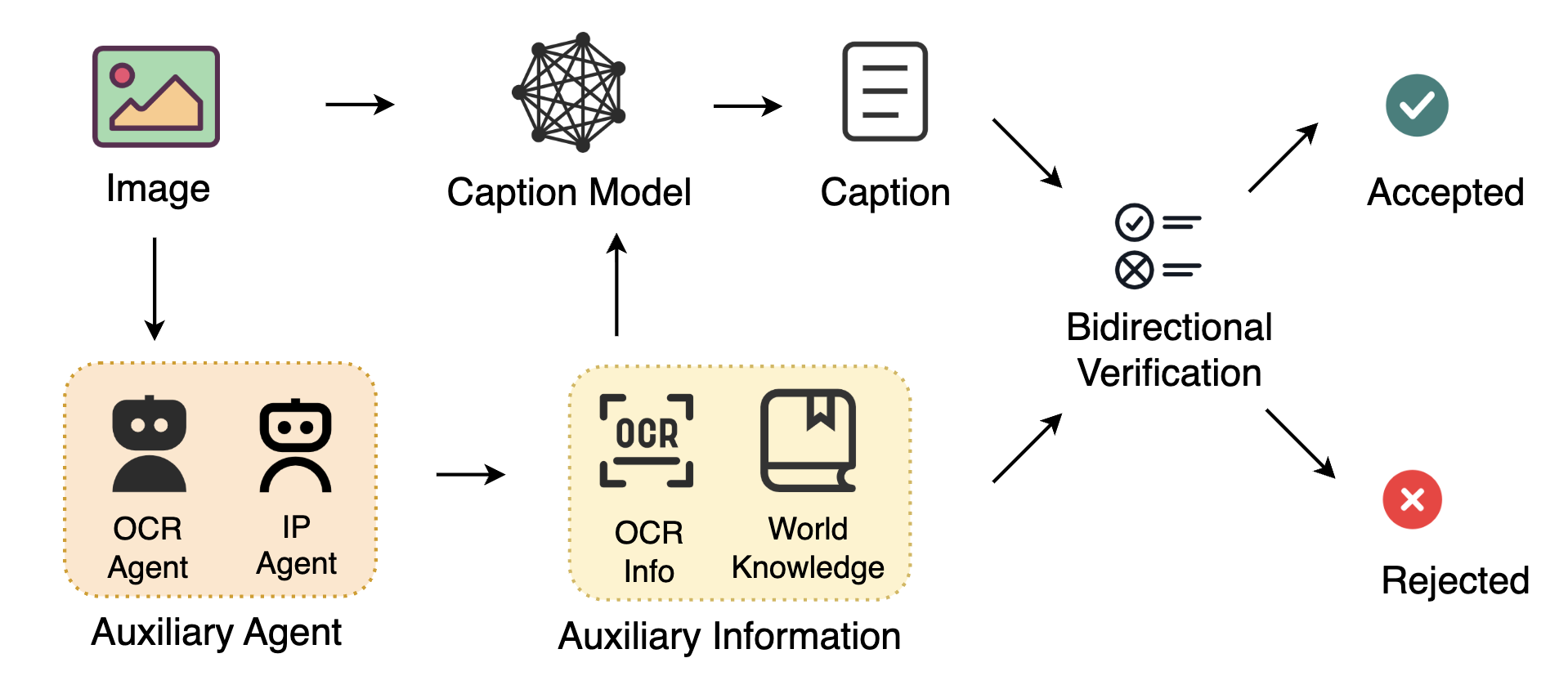}
    \caption{ Image Captioning Pipeline.}
    \label{fig:caption}
\end{figure}

To generate rich, controllable, and factually-grounded image descriptions, we propose a novel pipeline built upon three core components: (1) a hierarchical schema for structured image description, (2) a compositional synthesis strategy for diverse data augmentation, and (3) specialized agents for factual grounding, as illustrated in \figref{fig:caption} .

\textbf{Bilingual and Hierarchical Captioning Schema}. 
The foundation of our approach is a bilingual (English/Chinese) and hierarchical captioning schema that decomposes image content into multiple, well-defined semantic fields. This structured representation includes:
\begin{itemize}
\item \textbf{Descriptive Levels (Short to Extra-Long)}: Four tiers of narrative detail, from a concise summary to an exhaustive depiction of all foreground and background elements.
\item \textbf{Stylistic Attributes}: Fields for capturing the image's artistic style, cinematographic shot type, lighting, prevailing atmosphere, and composition.
\item \textbf{Factual Entities}: A dedicated field for named entities (IP), identifying specific characters, landmarks, brands, and artworks.
\end{itemize}

This hierarchical schema not only enables fine-grained control over the generative process, but also serves as the structural basis for our data synthesis engine.

\textbf{Compositional Caption Synthesis for Data Diversity}. 
To enhance model generalization and mitigate overfitting, we introduce Compositional Caption Synthesis, a dynamic data augmentation strategy that leverages our hierarchical schema. During training, we strategically sample and combine different fields to generate captions varying in both length and pattern, supporting bilingual (English/Chinese) outputs from about 30 words up to 1,000 words.

\textbf{Factual Grounding via Specialized Agents and Bidirectional Verification}. 
To overcome the limitations of standard Visual Language Models (VLMs) in recognizing dense in-image text and entities requiring world knowledge, we integrate two specialized agents to ground the descriptions in verifiable reality. An OCR Agent extracts in-image text, while a Named Entity (IP) Agent identifies real-world entities. This external knowledge is fed as auxiliary input to the caption model. Crucially, we establish a Bidirectional Verification Loop that cross-references the entities detected by the agents with the generated caption. Following this, only the samples that successfully pass the bidirectional verification are included in the final training dataset.

\textbf{Image Difference Captioning}. Additionally, for paired image data, we have developed an image difference captioner. This model takes a pair of images, their captions and corresponding two-frame video as input to generate a caption detailing the changes in the foreground and background, serving to simulate user-input editing instruction. The contextual information is important to enable the model to generate more accurate and detailed difference descriptions.

\subsection{Reasoning Dataset Construction}\label{sec:reasoning_data}

Our model is a powerful, natively multi-modal architecture endowed with robust reasoning and semantic understanding capabilities. A key contribution of our work is the elicitation of an automated \textit{Chain-of-Thought (CoT)} reasoning process for image generation, which is activated through fine-tuning on a small, specialized dataset. This process enables the model to autonomously execute a full pipeline: from interpreting an initial input prompt, to engaging in an intermediate "thinking" phase of conceptual refinement and rewriting, and finally to synthesizing the target image. To effectively unlock this latent ability, we constructed two specific types of training data: \textbf{(1)} Text-to-Text (\textbf{T2T}) reasoning data to bolster its logical inference. \textbf{(2)} Text-to-Text-to-Image (\textbf{T2TI}) reasoning data that explicitly models the entire workflow from abstract concepts to their visual manifestations. This training strategy empowers the model to achieve a seamless, automated, and coherent translation from high-level user intent to high-fidelity visual output.

\textbf{Text to Text.} To enhance the model's instruction following and logical-reasoning ability, we curated a diverse textual corpus of real-world image generation prompts. This corpus spans photorealistic rendering, artistic and stylistic renderings, UI and poster design tasks, knowledge-driven queries, and scientific or technical visualizations. By covering a broad spectrum of user intents, domains, and complexity levels, the model trained with T2T data can parse nuanced requirements, resolve ambiguities, and produce coherent, stepwise textual reasoning that faithfully maps instructions to precise image captions.

\textbf{Text to Text\&Image.} To improve end-to-end textual reasoning and visual fidelity, we developed the T2TI corpus, a high-quality, class-balanced image dataset. Images were filtered from the the pretraining dataset using aesthetic metrics and paired with their original short and long captions. We also compiled a collection of infographics from Wikipedia. For each image, we annotate a corresponding reasoning trace that refines goals and translates user intent into detailed visual specification. The images along with their captions, reasoning traces are used to improve the model's CoT image generation ability. 

\textbf{Text\&Image to Text\&Image.} To enhance the model's capability for complex image editing, we constructed the TI2TI corpus, a diverse dataset of editing pairs. The corpus consists of source images, complex editing instructions, and the corresponding ground-truth edited images. Crucially, for each instruction, we annotate a detailed editing trace. This trace deconstructs the user's complex instruction into a sequence of atomic operations. This collection of source images, instructions, editing traces, and final images is designed to train the model to perform step-by-step reasoning for image editing, thereby significantly improving its ability to follow complex and compositional user instructions.

\section{Model Design}\label{sec:model}

\subsection{Native Multimodal Model}
\begin{figure}[h]
    \hspace*{-0.75cm}
    \includegraphics[width=1.1\textwidth]{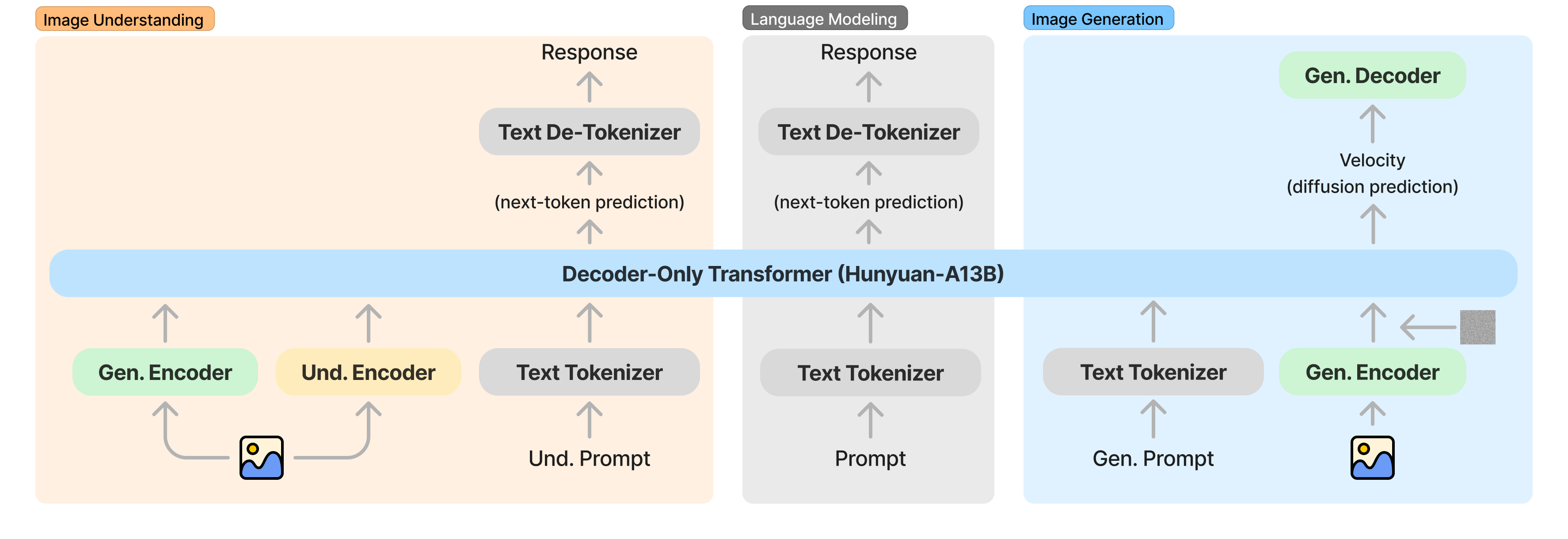}
    \vspace*{-0.5cm}
    \caption{Illustration of \nameofmethod{}.}
    \label{fig:model}
\end{figure}

We present a native multimodal model designed for unified understanding and generation across text and image modalities. As illustrated in \figref{fig:model}, the proposed architecture adopts a hybrid discrete-continuous modeling strategy: text tokens are modeled via autoregressive next-token prediction, while image tokens are modeled through a diffusion-based prediction framework~\cite{lipman2022flow}.
The system, designated as \nameofmethod{}, brings together language modeling, image understanding, and image generation in a cohesive framework to achieve unified multimodal modeling.

\subsubsection{Architecture}

\paragraph{Backbone.} The model backbone is constructed upon the Hunyuan-A13B~\cite{HunyuanA13B}, a decoder-only LLM over 80 billion total parameters. It employs an MoE configuration comprising 64 experts, with 8 experts activated per token, accompanied by one shared multi-layer perceptron (MLP). This design results in approximately 13 billion activated parameters per token, balancing expressive capacity with computational efficiency.

\paragraph{Text Tokenizer.} For textual input, we utilize the Hunyuan Tokenizer~\cite{HunyuanA13B}, extending its vocabulary with a set of custom special tokens tailored to support image generation and understanding tasks.

\paragraph{Image Encoder.} In the image generation pathway, we employ an internal VAE that projects raw pixel values into a 32-dimensional latent space with a downsampling factor of 16. Prior approaches, such as those in DiT-like~\cite{dit} architectures~\cite{sd3,hunyuandit,Flux,hunyuanvideo} typically combined an 8x downsampling VAE with an additional patchification layer that further reduced spatial resolution by a factor of 2. In contrast, we demonstrate that a single VAE with 16x downsampling offers a simpler and more effective alternative, yielding superior image generation quality. 

For conditioned image inputs, we introduce a dual-encoder strategy that concatenates latent features from the VAE with those from a vision encoder. This approach enables unified multimodal representation that supports both generation and understanding within a single sequence---a key different from previous unified models~\cite{show-o,janus-pro,bagel,mogao}, which often segregated visual features by task (e.g., using vision encoder features for understanding and VAE features for generation). Our method facilitates complex multimodal interactions---such as interleaved text dialogue, image generation, image understanding, and image-editing---within a continuous context, thereby eliminaing the need to switch between separate understanding and generation pipelines.

\paragraph{Projector.} We design two distinct projectors modules to align features from the dual image encoders into the transformer's latent space. Features from the VAE are projected using a timestep-modulated residual block~\cite{dit,resnet}, whereas features from the vision encoder are transformed via a two-layer MLP. Additionally, we incorporate timestep embedding into the sequence to enhance the conditioning of the diffusion process.

\begin{figure}[h]
    \centering
    \includegraphics[width=1.0\textwidth]{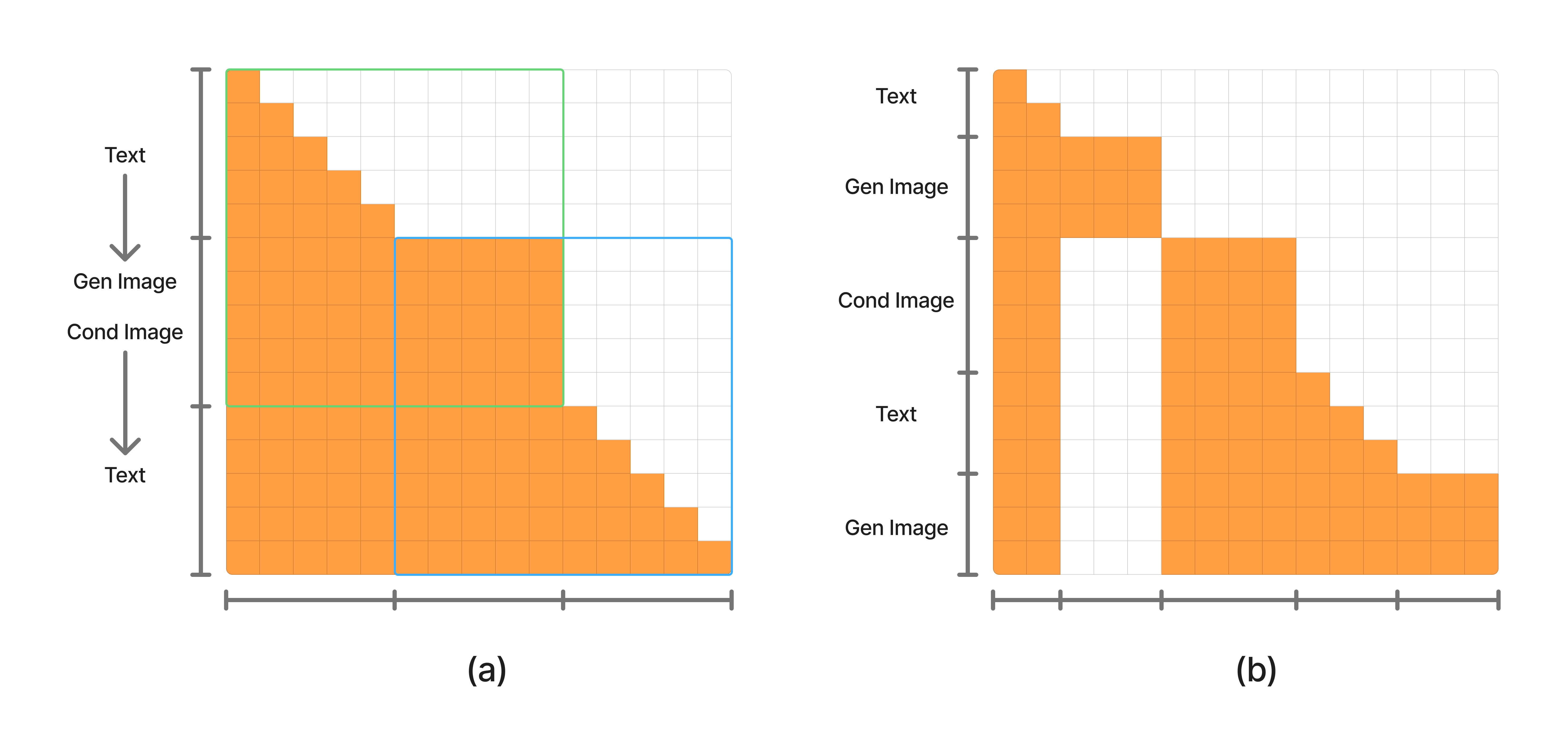}
    \vspace{-1.0cm}
    \caption{Two types of attention implementation.}
    \label{fig:attention}
\end{figure}

\subsubsection{Generalized Causal Attention} 
Causal attention is a fundamental component in LLMs for autoregressive text generation, as it ensures each token only attends to preceding tokens, thereby preserving the autoregressive property. In contrast, full attention is commonly employed in DiTs for image generation, allowing each image token to attention to all other tokens within the same image, which is beneficial for capturing global spatial dependencies. In our proposed native multimodal model, we integrate both attention types to handle heterogeneous data modalities effectively. Specifically, we introduce a \emph{Generalized Causal Attention} mechanism. Within this scheme, text tokens are restricted to attend only to previous multimodal tokens in the sequence. Image tokens, however, are permitted to attend to all previous multimodal tokens as well as all successive image tokens within the same image segment. This design respects the autoregressive geneartion nature of text while leveraging the global contextual capacity of full attention for image patches. 

As illustrated in \figref{fig:attention} we categorize the training attention mask into two distinct types based on the number of generated image segments (Gen Image), which correspond to the noised images being processed. In sequences where there are no Gen Images (as in image understanding tasks, indicated by the blue box in \figref{fig:attention}~(a)) or exactly one Gen Image (as in text-to-image tasks, green box in \figref{fig:attention}~(a)), the attention mask adheres strictly to the Generalized Causal Attention pattern defined above. However, when multiple Gen Images are present within a single training sequence (\figref{fig:attention}~(b)), a modification is necessary: any Gen Images appearing in the context must not be attended to by subsequent tokens in the sequence. This constraint introduces a ``hole'' (i.e., a region of mask attention) in the lower triangular part of the attention mask.

During inference, the input sequence never contains more than one simultaneous Gen Image. This is because once an image is generated, it is treated as a conditional image (Cond Image) for subsequent tokens in the sequence. Thus, the attention mask during inference consistently follows the cananical Generalized Causal Attention structure without requiring the additional masking needed during multi-gen-image training. This approach ensures causal consistency in generation while enabling effective multimodal learning.

\begin{figure}[h]
    \centering
    \includegraphics[width=1.0\textwidth]{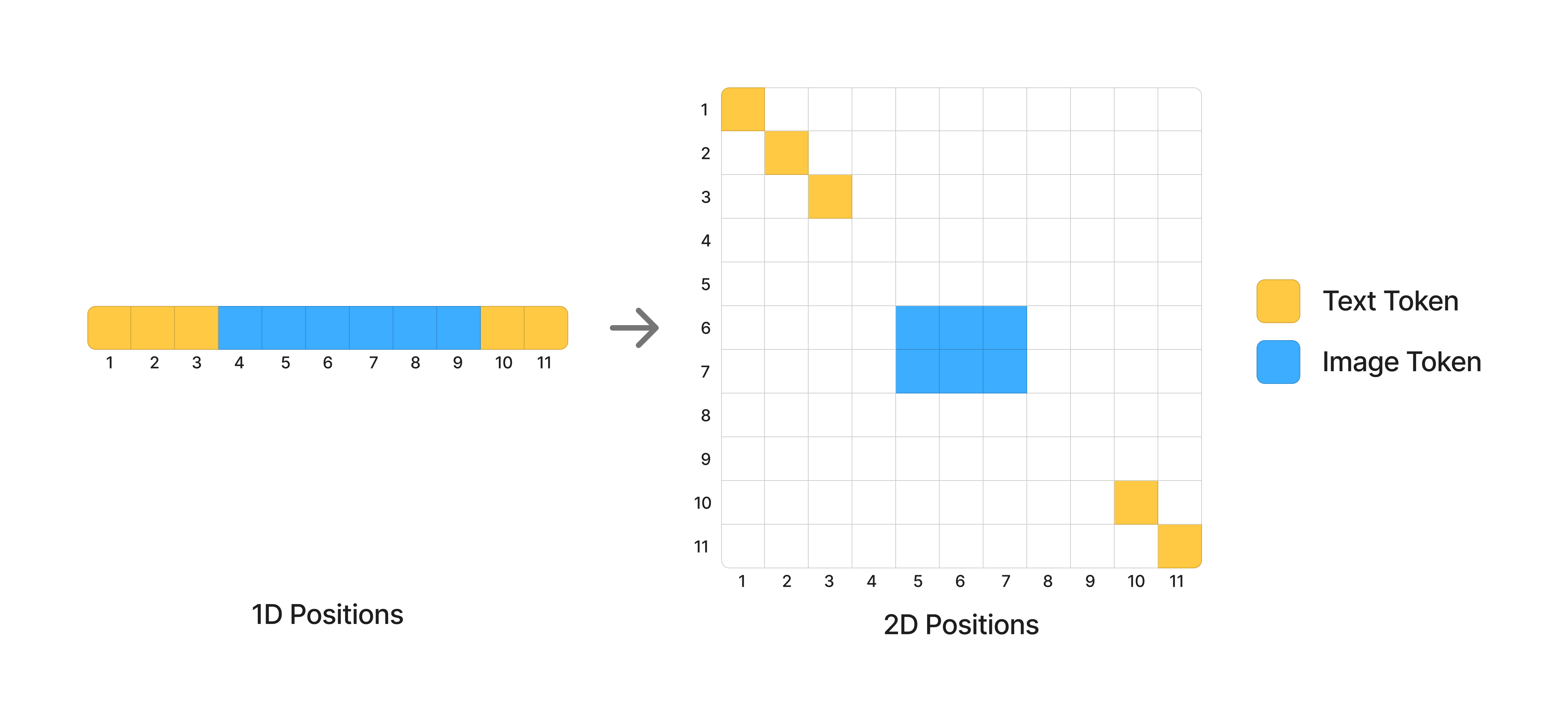}
    \vspace{-1.0cm}
    \caption{Illustration of the comparison between 1D RoPE and \emph{Generalized 2D RoPE} with backward compatibility. The text tokens in 1D RoPE has exactly the same positions as 2D RoPE. Intuitively, this generalization is implemented by reshape the 1D image positions into 2D positions and place it in the middle of two text sections.}
    \label{fig:rope_2d}
\end{figure}

\subsubsection{Position Embedding} Rotary Position Embedding (RoPE)~\cite{rope} is widely adopted in LLMs and DiTs due to its flexibility and scalability. In this work, we implement a \emph{Generalized 2D RoPE}, as proposed by Su.~\footnote{https://kexue.fm/archives/10352} This approach maintains backward compatibility with the pretrained LLM. Formally, for a one-dimensional text position index $n$, and a set of frequencies $\{\theta_0,\theta_1,\dots\}$, the position embedding is defined as $[\cos(n\theta_0),\cos(n\theta_1),\dots,\sin(n\theta_0),\sin(n\theta_1),\dots]$. We generalize this formulation to two-dimensional coordinates by interpreting the same embedding structure anisotrapically. For a position $(x,y)$, the embedding becomes $[\cos(x\theta_0),\cos(y\theta_1),\dots,\sin(x\theta_0),\sin(y\theta_1),\dots]$. 
As depicted in \figref{fig:rope_2d}, image tokens---which are reshaped from 1D to 2D---are assigned such generalized 2D position encodings, while text tokens retain standard 1D RoPE, and also can be viewed as 2D RoPE of diagnoal positions. This design ensures that in the absence of image tokens, the encoding reduces exactly to 1D RoPE, thereby preserving full compatibility with conventional text generation and minimizing disruptive effects on pre-trained linguistic capabilities.

In the training sequence incorporating multiple Gen Images (\figref{fig:attention}~(b)), the tokens following each Gen Image are assigned different positions in the training and inference sequences. To ensure positional consistency between training and inference, the position embeddings for these tokens are adjusted by shifting their token positions accordingly. This alignment is critical for maintaining the structural integrity of the sequence during model training and inference, as it mitigates potential discrepancies introduced by the variable placement of generated images. By explicitly aligning the positional encoding scheme across both phases, the model can more effectively generalize and preserve coherent contextual relationships within the sequence.

\subsubsection{Automatic Resolution}

DiT-like models typically require deterministic user input to specify the desired image size and aspect ratio. In the proposed native multimodal model, we introduce an automatic mode that allows the model to determine appropriate image shapes based on the context, which can be use prompt or conditional image tokens. Specifically, we extend the language model’s vocabulary with two groups of special tokens: one set represented as \{<img\_size\_256>, <img\_size\_512>, <img\_size\_768>, ...\}, and the other as \{<img\_ratio\_0>, <img\_ratio\_1>, <img\_ratio\_2>, ..., <img\_ratio\_32>\}. Each <img\_size\_*> token corresponds to an image resolution anchor, while each <img\_ratio\_*> token represents an aspect ratio ranging from 1:4 to 4:1. During training, the model learns to associate these shape tokens with the user inputs and previous conversations in the context, enabling it to predict appropriate size and ratio tokens according to the input context. Additionally, users can provide explicit cues—such as “3:4” or “vertical”—to guide the model toward generating a specific aspect ratio token. Based on the predicted size and ratio tokens, we can incorporate the 2D RoPE for the image tokens, enabling the model to generate images with the desired structural properties.

\section{Model Training}\label{sec:training}

\subsection{Pre-training}

Given the diversity of tasks involving heterogeneous token sequences in multimodal modeling, we design a flexible multi-task training framework capable of supporting large-scale training across numerous tasks and datasets, including text-to-image generation (T2I), language modeling (LM), multimodal understanding (MMU), interleaved text-image modeling (INTL) and reasoning (CoT).
Our pre-training process is organized into four progressive stages, in which training data are filtered from coarse to fine and image resolutions are increased gradually for VAE encoder and fixed for ViT encoder.
During training, the aspect ratio of images are preserved to enable the multi-resolution image generation. Detailed configurations for each stage are provided in \tabref{tab:training_stages}.

\begin{table}[h]
\centering
\caption{Training stages of the proposed native multimodal model. Resolution anchor denotes that the images are resized to the desired size while keeping the aspect ratio. We adopt progressive image resolution anchors for VAE and a fixed resolution anchor for vision encoder (ViT).}
\label{tab:training_stages}
\resizebox{\textwidth}{!}{
    \begin{tabular}{ccccc}
    \toprule
     Training Stage & VAE Reso. Anchor & ViT Reso. Anchor & Training Part & Task \\
     \midrule
     I & 256px & 512px & Transformer & T2I, LM, MMU \\
     II & 256px & 512px & ViT & MMU \\
     III & 512px & 512px & ViT, Transformer & T2I, LM, MMU, INTL \\ 
     IV & 1024px & 512px & ViT, Transformer & T2I, LM, MMU, INTL, CoT \\
     \bottomrule
    \end{tabular}
}
\end{table}

\paragraph{Progressive Training.} In the first stage, we train the Transformer backbone while keeping the ViT frozen. Three tasks are optimized simultaneously: text-to-image (T2I), language modeling (LM), and multimodal understanding (MMU), utilizing both text-image pairs and text-only data. This stage employs a low image resolution (256px) for VAE encoder and a large batch size training, enabling the model to learn from billions of images and align the latent representations of text and image modalities. During the second stage, the Transformer backbone remains frozen, while the ViT and its associated aligner module are fine-tuned using only MMU data to enhance visual understanding capabilities.
In Stage III, both the ViT and Transformer are jointly trained with images of higher resolution (greater than 512px). The dataset size is reduced to increase the proportion of high-quality images.
Interleaved text-image data, such as image editing and image-to-image data, are incorporated at this stage to enhance multimodal modeling capabilities.
In the final stage, training images are further constrained to a high-resolution subset, each with at least 1024 pixels on the shorter edge. Similarly, images used for the MMU task are limited to a high-resolution subset to enhance understanding capability. Although the input image size for the ViT encoder remains fixed at 512 pixels, we observe that high-resolution VAE features also contribute to improved model understanding.
In addition, reasoning data introduced in \secref{sec:reasoning_data} are incorporated at this stage to enable reasoning in multimodal modeling, particularly for Chain-of-Thoughts-based text-to-image generation.
Significantly, tokens of reasoning part are also modeled via autoregressive next-token prediction.

\paragraph{Instruction Tuning.}
After pre-training our native multimodal large language model, we perform instruction tuning specifically tailored for text-to-image generation.
At this stage, T2I, LM, and CoT data are formatted using instruction templates and jointly used to optimize the model.

\subsection{Post-training}

The post-training optimization of our model is a multi-stage process designed to systematically refine its generative capabilities. We first conduct SFT on a meticulously curated dataset of human-annotated examples. Following this, DPO~\cite{wallace2024diffusion} is implemented to effectively reduce visual artifacts and physical distortions. We then utilize MixGRPO~\cite{li2025mixgrpo} to enhance the critical aspects of text-image alignment, realism, and aesthetic appeal. The final refinement is achieved through the application of SRPO~\cite{shen2025directly} and a novel, in-house  Reward Distribution Alignment (ReDA) method, which together are instrumental in further elevating the realism and clarity of the generated images.

\paragraph{SFT.} For the SFT stage, we curate a high-quality text-to-image dataset spanning diverse domains, including landscapes, portraits, and OCR. This is further expanded into a comprehensive editing dataset by systematically pairing image types with specific editing operations, governed by stringent filters to ensure spatial consistency and identity preservation. To enhance the instruction-following capabilities of the model, we integrate sophisticated reasoning data. Finally, a multi-stage training strategy is employed, where subsequent phases progressively incorporate higher-fidelity samples to refine performance.

\paragraph{DPO.} We employ DPO to mitigate prevalent issues in text-to-image and editing tasks, such as structural distortion, sampling instability, and perceptible synthetic artifacts. Our DPO is structured into two distinct stages. In the first stage, we utilize an extensive corpus of paired data to bolster sampling stability and ensure consistency in editing. The second stage focuses on refining visual realism by leveraging a curated, high-quality subset of samples. This hierarchical training strategy effectively suppresses structural defects while significantly reducing visual artifacts and enhancing the overall aesthetic fidelity of the generated images.

\paragraph{MixGRPO.} MixGRPO is an efficient online reinforcement learning framework that extends GRPO to flow-based models through a hybrid ODE–SDE sampling strategy. We apply MixGRPO with proprietary reward models to optimize aesthetics (style, composition, lighting), mitigate distortions, and reduce artifacts. In addition, we refine advantage estimation to accelerate convergence, and demonstrate the scalability of MixGRPO to large-scale training regimes, achieving stronger alignment with human preferences.
To address a variety of image-to-image tasks, including image editing and identity (ID) preservation, we employ a balanced multi-task joint training strategy. This approach involves iteratively training multiple reward models, each targeting a different quality dimension. For the face ID preservation task, we specifically investigated and tailored a suitable reward model. Furthermore, we curated a challenging and diverse dataset of image-to-image pairs to guide model convergence more efficiently. Through these comprehensive explorations, we achieved significant improvements in the model's ability to preserve non-edited regions, maintain facial identity, and enhance overall image quality.

\paragraph{SRPO.} SRPO is a novel gradient-guided online reinforcement training strategy designed to enhance the realism and aesthetic quality of generated images. It directly injects a noise prior into the latent space features and then denoises it to a clean image in a single step. It selects the initial interval of the denoising trajectory for optimization, where the model has greater flexibility for improvement. By incorporating differentiable reward signals from both positive and negative text guidance, the model can efficiently align with human preferences and mitigate common issues in AI-generated images, such as oversaturation, incoherent lighting and colors, and poor skin texture. 

\paragraph{ReDA.} Reward Distribution Alignment optimizes the post-training phase by minimizing the divergence from a high-reward prior. We employ task-specific projectors to map generations into a compressed space, allowing for targeted optimization of metrics like identity consistency and realism. When reference images are provided, we introduce a transition-based objective: instead of comparing static features, we calculate the vector difference (transition) between the reference and the generated images. This formulation significantly boosts training efficiency. Finally, the algorithm is designed to be data-centric, effectively leveraging improvements in dataset scale and quality.

\subsection{Distillation}

To accelerate generation speed, we propose a novel distillation framework that scales MeanFlow~\cite{geng2025mean} for distilling the 80B-parameter unified multimodal model. Within this framework, we effectively mitigate the training instability issue, and further incorporate trajectory distribution alignment into the MeanFlow objective to enhance the few-step generation performance. Ultimately, we reduce the Number of Function Evaluations (NFE) to 4–8 while preserving competitive model performance.

\subsection{Pruning}

To reduce the deployment cost of HunyuanImage-3.0, we propose Tree-structured Mixed-policy Pruning (TMP), a post-distillation compression framework that combines a heuristic pruning strategy with hierarchical local mixed-policy distillation to support stable and aggressive model compression. Using TMP, we compress HunyuanImage-3.0 from 80B to 20B parameters (75\% reduction) while maintaining competitive generation quality. Combined with memory-efficient inference optimizations, the compressed model can be deployed on a single 24GB RTX 4090 GPU.

\section{Model Performance}\label{sec:performance}



\subsection{SSAE}
Modern Text-to-Image (T2I) generative models rely on standardized benchmarks like T2I-CompBench \cite{huang2023t2i} and GenEval \cite{ghosh2023geneval} to measure progress. However, these benchmarks exhibit limitations in comprehensiveness and reliability:

(1) Deficiencies in Prompt Design and Semantic Diversity: They often use short, formulaic structures (e.g., "a photo of a [object] with [attribute]"), failing to capture the complexity of real-world user instructions. This lack of diversity inadequately stress-tests model capabilities in parsing longer, intricate natural language descriptions involving multi-attribute composition, complex relational reasoning, and contextual logic.

(2) Over-reliance on Automated Metrics Misaligned with Human Judgment: Both benchmarks depend heavily on automatic metrics like CLIP Score for evaluating text-image alignment. However, these metrics are poor proxies for human assessment, as they may highly rate images with critical failures in spatial relationships (e.g., confusing "a boy under a bee" with "a bee under a boy") or precise attribute binding. This creates a disconnect between benchmark scores and human-perceived utility.

To address these issues, we propose a \textbf{s}tructured \textbf{s}emantic \textbf{a}lignment \textbf{e}valuation metric, \textit{abbr.}, \textbf{SSAE}. This intelligent metric leverages advanced LLMs and MLLMs for image-text alignment.

To resolve (1), we collect 500 diverse prompts and extract 3,500 key points using an LLM-based structured semantic point parser. Through in-context learning, points are categorized into 12 fine-grained fields: Nouns, main attributes and actions of primary and secondary subjects, other attributes of primary subjects, nouns and attributes of the scene, as well as camera shot, style, and composition. Another LLM then examines coherence between extracted points and original prompts, filters hallucinated points (e.g., objects or relationships inconsistent with prompts), and complements missing points, followed by human rectification. These points remain fixed during subsequent MLLM assessment to ensure stable and fair comparison across T2I models.

To resolve (2), an advanced MLLM with Chain-of-Thought reasoning scores model-generated images based on the prompts and pre-extracted key points, performing 0-1 matching. From this, we calculate both field-specific accuracy and two overall metrics: Mean Image Accuracy (mean accuracy of image-wise averaged scores) and Global Accuracy (averaged score across all key points in the dataset).

\begin{figure}[h]
    \centering
    \includegraphics[width=1.0\textwidth]{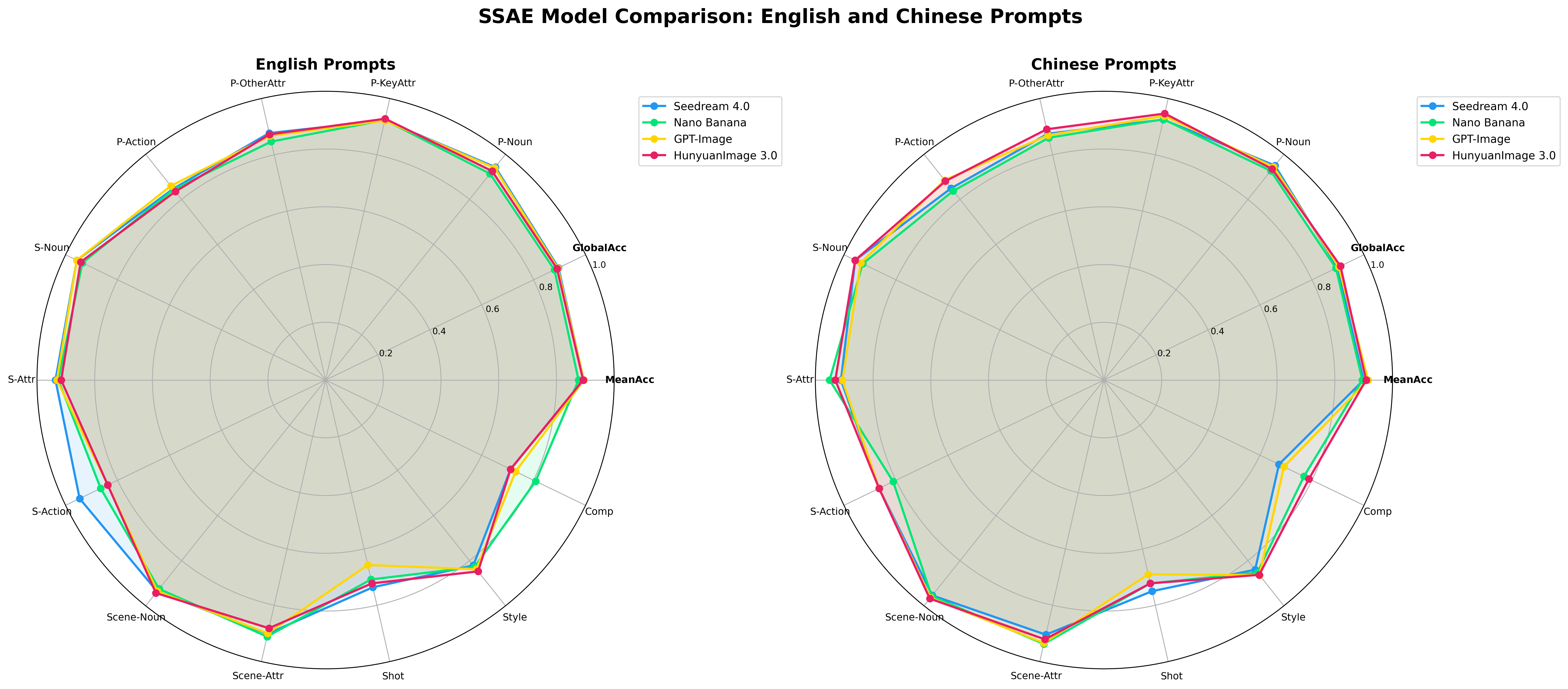}
    \caption{SSAE evaluation results.}
    \label{fig:ssae}
\end{figure}

Compared to recent benchmarks like DreamBench++ \cite{peng2024dreambench++}, which also uses MLLMs for human-like assessment, our benchmark features a more robust and comprehensive hierarchical semantic point parsing paradigm, taxonomy, and richer extensions (e.g., ranking mode beyond scoring).
The SSAE evaluation results for our method and competitors are shown in \figref{fig:ssae}.
As illustrated, \nameofmethod{} achieves performance on par with leading models in all fine-grained fields.

\subsection{GSB}
We adopt the GSB (Good/Same/Bad) evaluation method, which is commonly used to assess the relative performance of two models from an overall image perception perspective.
In practice, we carefully construct 1,000 text prompts to cover a blanced scenarios, and generate an equal number of image samples for each model in a single run.
For fairness, inference is performed only once for each prompt, without any cherry-picking of results.
All other models are evaluated under their default settings.
The evaluation is conducted by over 100 professional evaluators.

\begin{figure}[h]
    \centering
    \includegraphics[width=1.0\textwidth]{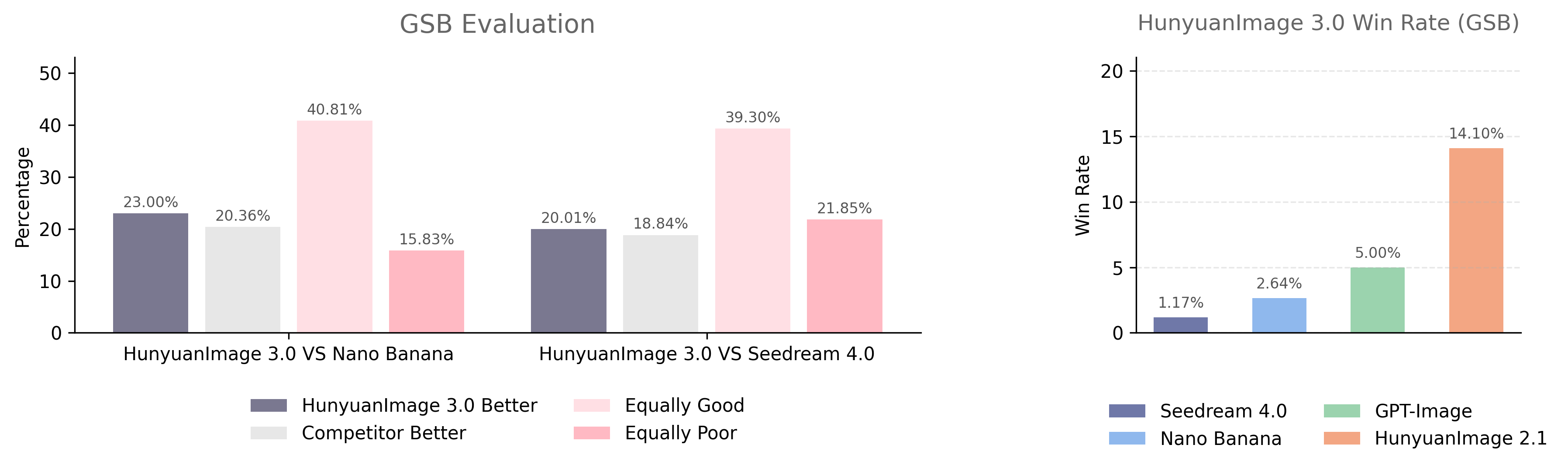}
    \caption{GSB evaluation results.}
    \label{fig:gsb}
\end{figure}

\figref{fig:gsb} presents the GSB evaluation results.
As shown, \nameofmethod{} achieves a relative win rate of 14.10\% compared to HunyuanImage 2.1, which was previously the best open-source model, thereby establishing \nameofmethod{} as the most powerful open-source text-to-image model to date.
Moreover, \nameofmethod{} achieves relative win rates of 1.17\%, 2.64\%, and 5.00\% compared to Seedream 4.0, Nano Banana, and GPT-Image, respectively.
These results demonstrate that \nameofmethod{}, as an open-source model, has reached a level of image generation quality comparable to leading closed-source commercial models.

\begin{figure}[h]
    \centering
    \begin{subfigure}[b]{0.47\textwidth}
        \centering
        \includegraphics[width=\textwidth]{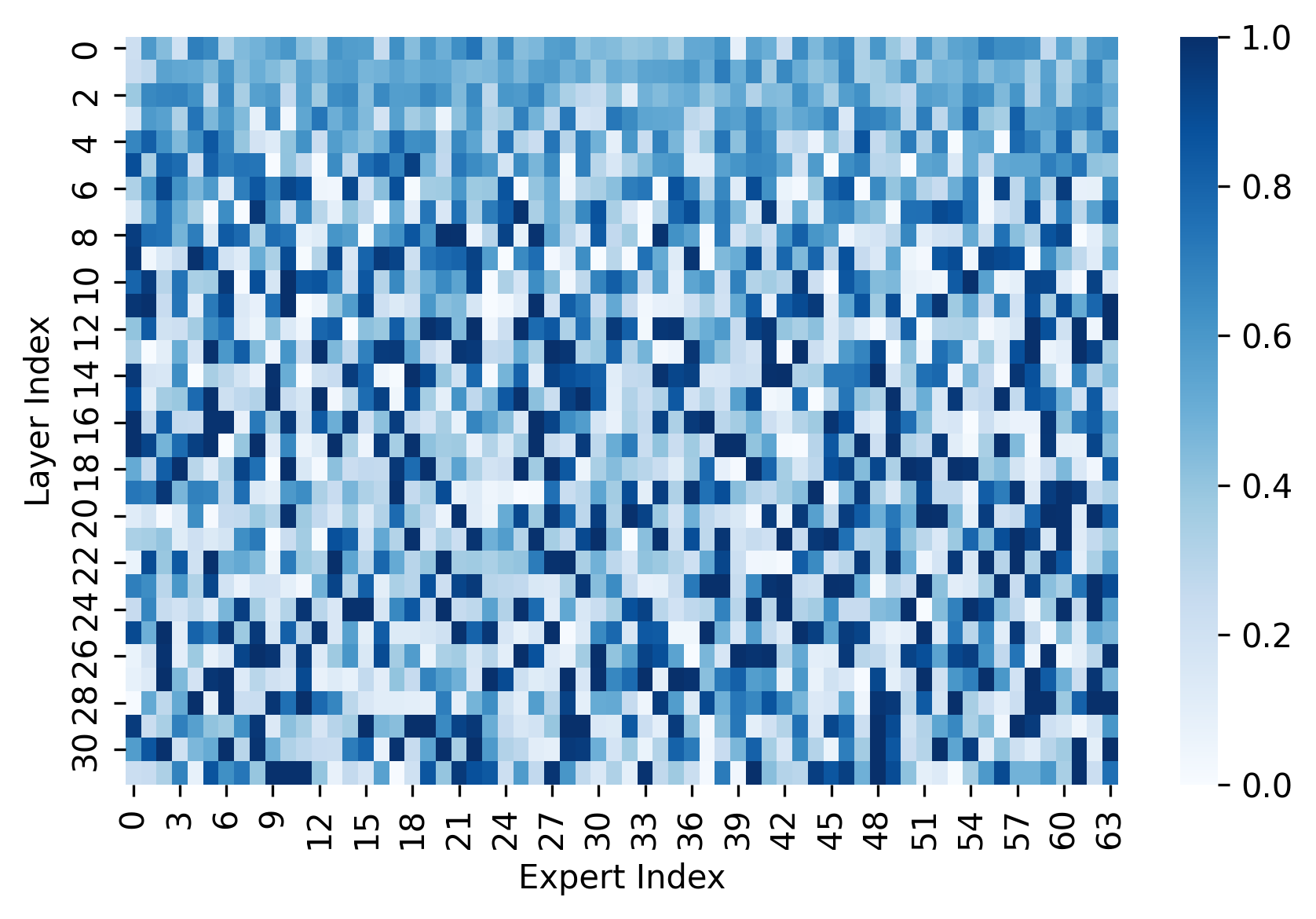} 
    \end{subfigure}
    \begin{subfigure}[b]{0.49\textwidth}
        \centering
        \includegraphics[width=\textwidth]{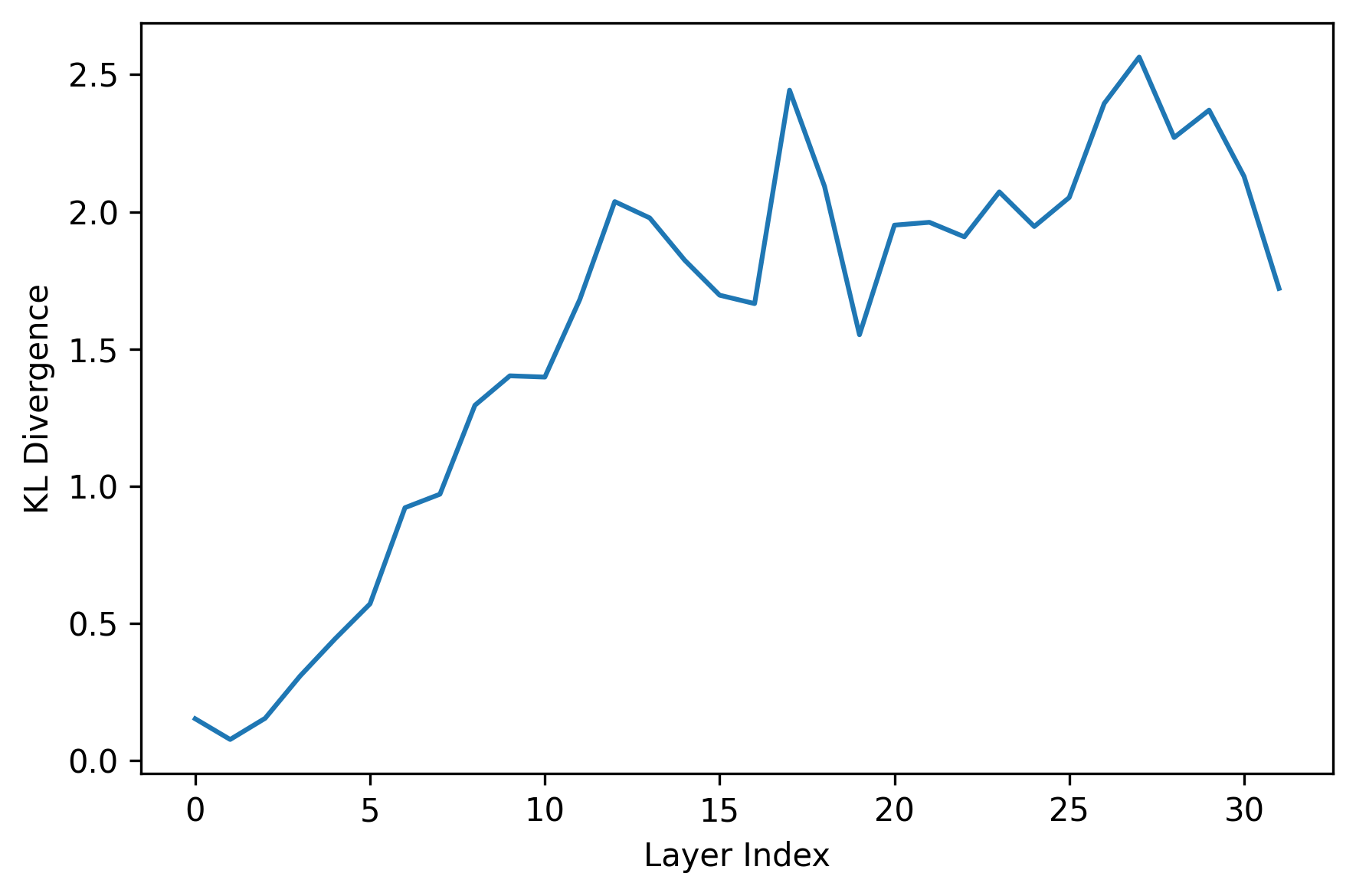} 
    \end{subfigure}
    \caption{Left: Heatmap of $\frac{\frac{v_{ij}}{\sum_{j} v_{ij}}}{\frac{v_{ij}}{\sum_{j} v_{ij}} + \frac{t_{ij}}{\sum_{j} t_{ij}}}$, where $v_{ij}$ and $t_{ij}$ denote the image-token and text-token activation counts, respectively, for the $j$-th expert in the $i$-th layer. The darker an expert appears, the more specialized it is to image tokens. Right: KL divergence between $\{\frac{v_{ik}}{\sum_{j} v_{ij}}\}_{k=0}^{64}$ and $\{\frac{t_{ik}}{\sum_{j} t_{ij}}\}_{k=0}^{64}$ for all experts of each layer. As the layer goes deeper, the KL divergence increases and the expert activation distributions become more dispersed across modalities.}
    \label{fig:expert_analysis}
\end{figure}

\subsection{Discovery}
\subsubsection{Expert Activation Analysis}

We analyse how experts are activated by tokens of different modals in multimodal MoE model.
We ramdomly select 1,000 prompts to perform text-to-image generation and conduct statistical analysis on experts of each layer using our pre-trained model.
\figref{fig:expert_analysis} demonstrates an expert modal preference heatmap and a KL divergence tendency between image- and text-activated-expert distribution for experts of each layer.
Both figures imply that the experts become increasingly specialized in individual modalities.
This suggests that MoE may enhance multimodal modeling by dispersing responsibilities for different modalities among specialized experts.

\section{Conclusion}
In this report, we present \nameofmethod{}, a native multimodal model that unifies multimodal understanding and generation within an autoregressive framework.
We begin with a pre-trained MoE LLM and extend it to support both image understanding and generation.
With large-scale pre-training on diverse and carefully curated multimodal data, our model demonstrates robust capabilities in both image understanding and generation.
Thanks to the LLM-based framework, we incorporate native Chain-of-Thought training and inference, which improves the multimodal performance significantly.
Building upon the pre-trained model, we perform fine-tuning and post-training specifically for image generation, and make the resulting model publicly available.
\nameofmethod{} exhibits strong capabilities in prompt-following, reasoning, concept generalization, and text rendering for text-to-image generation.
Results from both automatic and human evaluations on text-image alignment and visual quality indicate that \nameofmethod{} rivals existing state-of-the-art models.
While this release only includes the text-to-image ability, training for image-to-image tasks is ongoing, and this capability will be released in the near future.

\clearpage
\section{Project Contributors}

\begin{itemize}
  \item \textbf{Project Sponsors:} Jie Jiang, Liefeng Bo, Peng Chen, Yuhong Liu
    \item \textbf{Project Leader:} Zhao Zhong
    \item \textbf{Core Contributors:} (Authors in bolds are project leaders, others are listed alphabetically)
        \begin{itemize}
            \item \textbf{Captioner \& Data:} \textbf{Xin Li}, Duojun Huang, Xinchi Deng, Xuefei Zhe
            \item \textbf{VAE \& Model Accleration:} \textbf{Songtao Liu}, Changlin Li, Jianbing Wu, Peizhen Zhang, Yang Li
            \item \textbf{Algorithm \& Pretraining:} \textbf{Miles Yang}, Fanbin Lu, Jian-Wei Zhang, Qixun Wang, Shi-Xue Zhang, Yiji Cheng, Zijian Zhang
            \item \textbf{Post Training:} \textbf{Lucas Wang}, \textbf{Chunyu Wang}, Hangting Chen, Hao Wen, Junzhe Li, Lucaz Liu, Ringo Ling, Tao Zhang, Xiangwei Shen, Xindi Yang, Yingfang Zhang, Ying Dong, Yixuan Shi, Yutao Cui, Zheng Yuan, Zhengkai Jiang, Zheyuan Liu, Zhimin Li
        \end{itemize}
    \item \textbf{Contributors (Listed alphabetically):} Bing Wu, Chao Zhang, Donghao Li, Fang Yang, Jiale Tao, Jianchen Zhu, Jiaxin Lin, Jingmiao Yu, Kai Wang, Kipper Gong, Lei Wang, Linqing Wang, Pengfei Wan, Penghao Zhao, Qi Tian, Qinglin Lu, Senhao Xie, Shu Liu, Shuang Chen, Siyu Cao, Tiankai Hang, Tianpeng Gu, Weigang Zhang, Weijie Kong, Weiyan Wang, Xiaofeng Yang, Xiusen Gu, Xuan Yang, Yangyu Tao, Yanxin Long, Yepeng Zhang, Yu Liu, Yuanbo Peng, Yue Wu, Yuyang Peng, Zhantao Yang, Zhenxi Li, Zhiyuan Zhao, Zihao Zhang
\end{itemize}

\clearpage
{
\bibliographystyle{unsrtnat}
\bibliography{reference}
}

\end{document}